\documentclass[runningheads]{llncs}

\usepackage{eccv}

\usepackage{eccvabbrv}
\usepackage[utf8]{inputenc}
\usepackage{graphicx}
\usepackage{booktabs}
\usepackage{tabularx}
\usepackage{makecell}
\usepackage{multirow}
\usepackage{amsmath} 
\usepackage{siunitx}

\usepackage[accsupp]{axessibility} 
\usepackage[table]{xcolor}
\definecolor{rankone}{RGB}{255,210,210}   
\definecolor{ranktwo}{RGB}{173,224,235}   
\definecolor{rankthree}{RGB}{255,245,180} 

\newcommand{\best}[1]{\cellcolor{rankone}\textbf{#1}}
\newcommand{\second}[1]{\cellcolor{ranktwo}#1}
\newcommand{\third}[1]{\cellcolor{rankthree}#1}
\newcommand{\bestcap}[1]{\colorbox{rankone}{\textbf{#1}}}
\newcommand{\secondcap}[1]{\colorbox{ranktwo}{#1}}
\newcommand{\thirdcap}[1]{\colorbox{rankthree}{#1}}

\newcommand{\labw}{0.03\linewidth}   
\newcommand{\imgw}{0.13\linewidth}  

\newcommand{\img}[1]{%
\includegraphics[width=3cm,height=1.5cm,keepaspectratio]{#1}}

\usepackage{hyperref}

\usepackage{orcidlink}

\begin{document}

\title{Gaussian Volumetric Representation for Efficient Shear-Warp Visualization} 

\titlerunning{Gaussian Volumetric Shear-Warp Visualization}

\author{
Mayuri Mathur\inst{}\orcidlink{0009-0001-3456-7318}
\and
Ojaswa Sharma\inst{}\orcidlink{0000-0002-9902-1367}
}

\authorrunning{M. Mathur and O. Sharma}

\institute{
Indraprastha Institute of Information Technology Delhi, New Delhi, India\\
\email{mayurim@iiitd.ac.in, ojaswa@iiitd.ac.in}\\
\url{https://graphics-research-group.github.io/Gaussian-Volumetric-Representation/}
}

\maketitle

\begin{abstract}
Medical image visualization requires volumetric rendering algorithms that preserve anatomical fidelity while maintaining high rendering speeds. To address the high computational cost of large volumetric datasets, we propose a Gaussian-based volumetric representation for efficient visualization of dense medical volumes without compromising structural and radiometric details.
We optimize the proposed representation using Monte Carlo volumetric estimation, which enables training on a highly sparse subset of voxels while maintaining consistency with the dense volumetric objective. In addition, we introduce a curriculum learning strategy that progressively incorporates structured slice-based sampling during training. Sparse voxel samples provide an early global coverage of the volume, while slice samples capture spatially correlated regions that aid geometric structure and texture continuity. This combination enables the Gaussian representation to learn anatomical details of various structures and corresponding textures from sparse supervision while significantly reducing the computational cost associated with dense voxel processing. The learned representation supports slice-based rendering methods such as shear–warp volume rendering, enabling efficient visualization of multimodal medical datasets including MRI and Cryosection volumes while preserving anatomical structures. Using sparse supervision, our method achieves up to 43.86 FPS rendering with a compression ratio of 11.31:1. 
  \keywords{Gaussian representation \and Volume representation \and Monte Carlo estimation \and Curriculum Learning}
\end{abstract}

\section{Introduction}
\label{sec:intro}
Medical volumetric data like MRI, CT, and Cryosection data are high-dimensional datasets and resource-hungry during real-time visualization. The ray marching~\cite{kim2021image,devkota2022deep} volumetric visualization technique renders each pixel by sampling and interpolating along a camera ray through the volume, making it computationally expensive for high-resolution views and large volumetric datasets.

Current methods encompassing the domain of volumetric representation and compression comprise conventional\cite{kolda2009tensor,ahmed1974discrete,meagher1982geometric} and learning based techniques\cite{mildenhall2021nerf,10.1007/978-3-319-24574-4_28,vaswani2017attention}.
Conventional methods such as tensor decomposition\cite{kolda2009tensor}, discrete cosine transform\cite{ahmed1974discrete}, and octree-based\cite{meagher1982geometric} compression aim to reduce storage requirements of volumetric data by approximating the voxel grid using compact representations. While these methods provide efficient compression, they primarily operate on voxel-level signal approximations and do not explicitly model the volumetric structure required for rendering operations. As a result, these approaches are not well suited for applications that require flexible reconstruction of volumetric signals under varying rendering configurations, particularly for real-time visualization of large medical volumes.

Learning-based techniques such as feedforward networks\cite{mildenhall2021nerf}, convolutional neural networks (CNNs)\cite{cciccek20163d}, and Transformers\cite{wang2021multi} have been used to represent volumetric data through learned embeddings. Although these approaches can capture fine details, modeling large volumes often requires a large number of parameters, making them computationally expensive. Optimization-based rendering techniques \cite{duan2020curriculum,10.1145/3341156} have also been proposed for real-time reconstruction; however, most of these methods are surface-based and focus on reconstructing object boundaries rather than modeling the full volumetric interior. As a result, they fail to capture the fine cross-sectional structures present in dense volumetric datasets such as MRI and Cryosection.

Efficient learning of volumetric representations therefore, requires training strategies that approximate the dense voxel objective while minimizing computational cost. Gaussian splatting~\cite{3dGS, xu2025freesplatter,tang2025ivr} is primarily optimized for surface-like scene representations, and therefore struggles to directly model voxel-based medical volumes where meaningful information is distributed across internal cross-sectional structures rather than only on visible surfaces.
To address these limitations, we propose a Gaussian-based volumetric representation that learns a compact approximation of dense medical volumes. Inspired by the recent success of Gaussian splatting in real-time rendering, we extend this representation to volumetric medical data for efficient reconstruction and visualization. We optimize the proposed representation using a curriculum learning strategy that combines sparse voxel sampling and structured slice-based sampling during training. Sparse voxels provide global coverage of the volume, while slice samples capture spatially correlated regions that encode geometric structure and texture continuity. This combination enables the Gaussian representation to efficiently learn both global volumetric consistency and local anatomical details. 
Since shear-warp rendering operates on stacks of 2D slices, incorporating slice-based supervision during training naturally aligns the learned representation with the rendering process.

The proposed framework is applicable to multimodal volumetric datasets including MRI and Cryosection volumes. By learning a compact Gaussian representation of organ structures, the method enables efficient reconstruction and supports real-time selective visualization of individual organs.
Our contributions can be summarized as follows:
\begin{itemize}
\item A Gaussian-based volumetric representation that approximates a dense medical volume using a sparse set of learnable Gaussian kernels.
\item A training strategy that utilizes Monte Carlo voxel sampling for efficient optimization of the volumetric representation from sparse supervision.
\item A curriculum learning-based approach to training Gaussians using a mix of volumetric and planar slice-based sampling strategies. Incorporating slice samples provides spatially correlated supervision that improves reconstruction of anatomical structures compared to isolated voxel samples.
\item A fast GPU-based Gaussian volume renderer based on the Shear-warp algorithm for real-time visualization.
\end{itemize}

Furthermore, the learned representation naturally supports slice-based rendering techniques such as shear–warp volume rendering, enabling high frame-rate visualization of large volumetric datasets.

\section{Related Work}

\subsubsection{Learning-based methods in volumetric representation:}

Learning-based approaches have been explored for representing volumetric data using continuous implicit functions. Neural Radiance Fields (NeRF) \cite{10.1145/3503250,tang2023able,corona2022mednerf,khojasteh2025mis,ruckert2022neat,zha2022naf} approximate a volumetric function that maps 3D spatial coordinates and viewing directions to color and density values using a multilayer perceptron (MLP). Rendering a single image requires evaluating the network multiple times along each camera ray, making the rendering process computationally expensive due to repeated MLP inference and dense ray sampling.

Although NeRF can capture detailed volumetric signals, the use of large neural networks and dense sampling along rays leads to high computational overhead during both training and rendering. For high-resolution volumetric datasets such as medical scans, the number of network evaluations required to reconstruct fine spatial structures can become prohibitively expensive.

CNNs\cite{10.1007/978-3-319-46478-7_31,kim2021image,devkota2022deep,lyu2021ultrasound} have also been used to learn volumetric representations by operating directly on voxel grids. While CNN-based models can capture spatial context through convolutional filters, they often require dense volumetric inputs and large memory footprints when applied to high-resolution volumes.

More recently, Transformer\cite{vaswani2017attention,lin2023vision,liang2023retr,su2024dt,chen2022transformers,chang2026gat,cong2023enhancing,zhong2024cvt} models have been explored for volumetric representation learning by modeling long-range spatial dependencies across volumetric data. Although transformers can capture global context effectively, their computational complexity grows rapidly with the size of the input volume, making them challenging to scale for large medical datasets.
GNT-MOVE~\cite{cong2023enhancing} extends NeRF-based novel-view synthesis using Transformer-based view aggregation. CVT-xRF \cite{zhong2024cvt} is a NeRF-based technique that uses Transformers to build a more consistent 3D radiance field from multiple input views. It applies cross-view reasoning, where the Transformer exchanges and aggregates features across images of the same scene, assigning higher importance to the views that provide more useful information for reconstruction.
These limitations motivate the need for compressed volumetric representations that can efficiently capture volumetric structure while reducing the computational cost associated with dense neural representations.

\subsubsection{Surface rendering with Gaussians:}
Surface-oriented Gaussian Splatting methods have been widely explored for novel-view synthesis and visualization \cite{3dGS,liang2024analytic, tang2025ivr, kleinbeck2025multi, li2025clipgs, condor2025don, xu2025freesplatter}.
Kerbl et al. \cite{3dGS} introduced 3D Gaussian Splatting, where a scene is represented as a set of anisotropic Gaussian primitives parameterized by position, covariance, color, and opacity. These Gaussians are projected onto the image plane using a visibility-aware rasterization procedure, enabling real-time novel-view synthesis with high visual fidelity. Liang et al. \cite{liang2024analytic} proposed Analytic Splatting to reduce aliasing artifacts observed in Gaussian Splatting. Their method analytically integrates the Gaussian contribution across the pixel footprint, producing smoother renderings and preserving fine structural details. FreeSplatter~\cite{xu2025freesplatter} is a Gaussian Splatting framework that predicts 3D Gaussian primitives and camera parameters from uncalibrated sparse view images using a Transformer-based multi-view feature exchange technique. The Transformer exchanges and aggregates information across multiple input views, helping infer consistent 3D structures from sparse observations. Tang et al. \cite{tang2025ivr} introduced iVR-GS for interactive visualization of volumetric medical data. The method groups Gaussian models associated with different transfer functions to enable interactive exploration and editing of volumetric structures. Kleinbeck et al. \cite{kleinbeck2025multi} proposed a layered Gaussian representation for immersive visualization of anatomical structures. Each Gaussian layer corresponds to a specific anatomical component and is trained sequentially to reduce overlap between structures. ClipGS \cite{li2025clipgs} extends Gaussian splatting for interactive visualization of medical scans by introducing clipping operations that allow users to explore internal structures through dynamically adjustable slicing planes. Recent work \cite{condor2025don} reformulates Gaussian primitives as volumetric density functions and performs ray-based integration through the Gaussian field to solve the radiative transfer equation. This enables physically consistent volumetric rendering but introduces additional computational cost compared to rasterized splatting methods.In contrast to surface-oriented Gaussian splatting techniques, our work focuses on learning a Gaussian representation that models the interior volumetric structure of medical datasets while enabling efficient reconstruction and visualization.

\section{Method}

\begin{figure}[!ht]
    \centering
    \includegraphics[width=1\textwidth,height=0.6\linewidth]{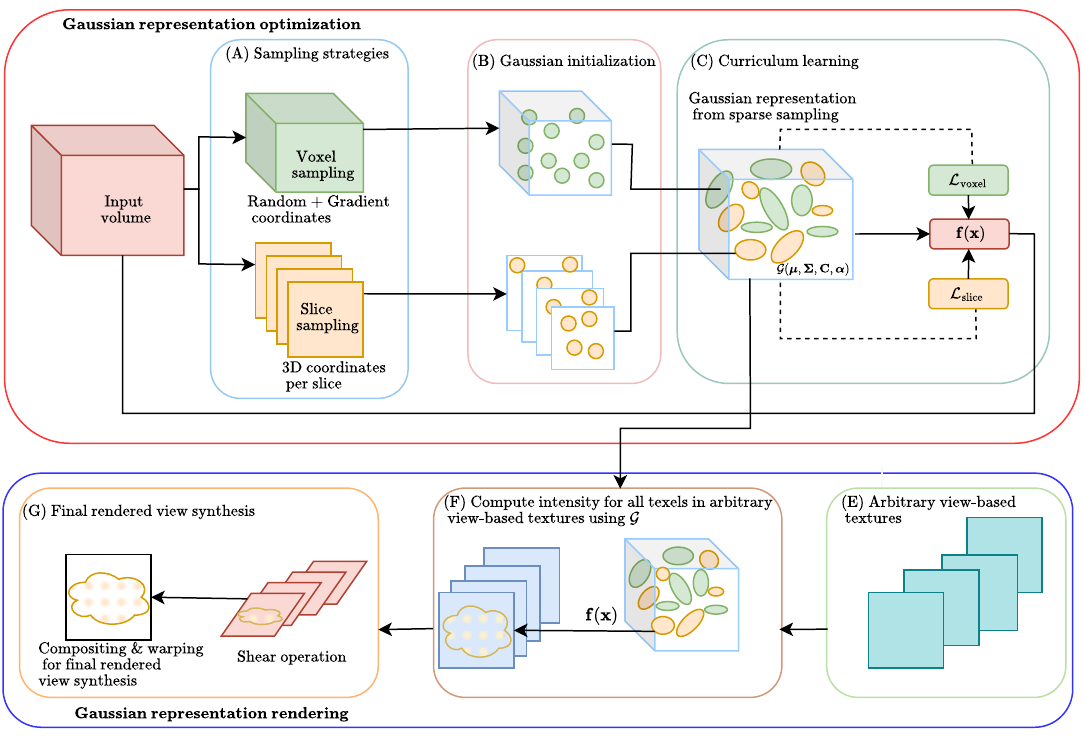}
    \caption{Overview of the proposed Gaussian-based volumetric representation for real-time rendering. Sparse voxel and slice samples are used under a curriculum sampling strategy to optimize a continuous Gaussian volumetric representation, which is subsequently used to generate arbitrary view-based slice textures for shear–warp volume rendering.}
    \label{fig:archi}
\end{figure}

As shown in Fig.~\ref{fig:archi}, our method represents volumetric data using a set of Gaussian kernels that approximate the underlying continuous volumetric signal. The Gaussian representation enables reconstruction of the entire voxel grid while significantly reducing the number of primitives required for its representation. The reconstructed color at any spatial location is obtained by evaluating the Gaussian mixture. To efficiently optimize this representation, we initialize Gaussians from a sparse set of voxels and update their parameters using sampled supervision. Instead of computing the reconstruction loss over the full voxel grid, we use a Monte Carlo volumetric estimator\cite{luengo2020survey} to approximate the dense objective using a subset of voxels while preserving consistency with the full-volume optimization. We further improve the learning of anatomical structures by combining sparse voxel sampling with slice-based sampling by progressively training these sampling strategies using Curriculum learning\cite{wang2019dynamic}. In each training iteration, a plane with arbitrary orientation is sampled, and sparse pixels are selected on the plane. Each pixel corresponds to a three-dimensional voxel coordinate, allowing the Gaussian representation to learn spatial continuity across neighboring regions.
This sampling strategy aligns with shear–warp volume rendering, which generates images from stacks of two-dimensional slices extracted from a three-dimensional volume. In the following sections, we first present our Gaussian representation designed for the shear–warp renderer, followed by an analysis of the performance improvements achieved by the proposed method.

\subsection{Gaussian representation}
\label{sec:gauss_repre}
Consider representing a given voxel grid with a set of Gaussian kernels\cite{3dGS} estimated from $m$ voxel samples. We initialize one Gaussian kernel centered at each sampled voxel position and define the set $\mathcal{G} = \{G_1, G_2, \ldots, G_m\}$ of $m$ Gaussians. Each Gaussian has associated learnable parameters:  mean $\boldsymbol{\mu} \in \mathbb{R}^3$, covariance matrix $\boldsymbol{\Sigma} \in \mathbb{R}^{3\times3}$, RGB color $\mathbf{C} \in \mathbb{R}^3$, and opacity $\alpha\in \mathbb{R}$ that are estimated during optimization. The Gaussian mean $\boldsymbol{\mu}$ determines the spatial location of the colored blob, while the covariance matrix $\boldsymbol{\Sigma}$ controls the spatial spread and orientation of the color contribution. Although each Gaussian has a constant color parameter $\mathbf{C}$, the Gaussian weighting function formulates its spatial influence, producing a continuous volumetric color field when multiple Gaussians are combined. This Gaussian representation defines a continuous volumetric field $\mathbf{f}$ of colors computed as the weighted contribution of all Gaussian kernels in the mixture
\begin{align}
\mathbf{f}(\mathbf{x}) =
\sum_{i=1}^{m}
\alpha_i \mathbf{C}_i
\exp\!\left(
-\frac{1}{2}
(\mathbf{x} - \boldsymbol{\mu}_i)^{\top}
\boldsymbol{\Sigma}_i^{-1}
(\mathbf{x} - \boldsymbol{\mu}_i)
\right).
\end{align}

\subsection{Monte Carlo volumetric estimation}
\label{sec:mce}
Let $\Omega$ denote the set of all $n$ voxels in the volume and $\mathbf{x}$ be the 3D  coordinate of a voxel. Our reconstruction objective is to minimize the voxel-wise mean squared error over the entire volume $V$
\begin{align}
L_{\text{dense}} = 
\frac{1}{n}
\sum_{\mathbf{x} \in \Omega}
\ell(\mathbf{x};\Theta),
\qquad
\ell(\mathbf{x};\Theta)
=
\|V_\Theta(\mathbf{x}) - V(\mathbf{x})\|_2^2,
\end{align}
where $V(\mathbf{x})$ is the ground-truth voxel color at coordinate $\mathbf{x}$ and $V_\Theta$ is the Gaussian-based continuous volume representation parameterized by $\Theta$. Optimizing the Gaussian volumetric representation using all voxels of a volume is computationally expensive for high-resolution grids, since every iteration requires computing the predicted color and gradient for each voxel that incurs significant memory and computational overhead. Therefore, directly minimizing the dense loss $L_{\text{dense}}$ is impractical for large volumetric datasets where grid sizes of $512^3$ are common.

A common alternative is to compute the loss using a subset of voxels sampled from the volume. One naive approach is to randomly sample a set of voxel positions in each epoch and compute the reconstruction loss only on these samples. That is, randomly sample $m \ll n$ voxel positions from a sampling distribution $q(\mathbf{x})$ defined over $\Omega$. In practice, this sampled voxel set corresponds to the sparse supervision set $\mathcal{M}$ introduced in Section~\ref{sec:mce}. The resulting loss can then be estimated as
\begin{align}
\hat L_{\text{naive}}
=
\frac{1}{m}
\sum_{i=1}^{m}
\ell(\mathbf{x}_i;\Theta),
\qquad \mathbf{x}_i \sim q(\mathbf{x}).
\end{align}
This estimator matches the dense objective when the sampling distribution is uniform, i.e., $q(\mathbf{x}) = \frac{1}{n}$. While uniform sampling produces an unbiased estimate, it is often inefficient for volumetric data since large regions may contain redundant or low-information voxels. In practice, it is beneficial to sample informative voxels more frequently, such as those with large reconstruction error or high spatial gradients. Therefore, such random subsampling does not guarantee that the estimated loss corresponds to the dense loss, particularly when the sampling distribution is non-uniform or when informative regions of the volume are sampled more frequently than others. In these cases, the optimization effectively minimizes a modified objective that depends on the sampling distribution rather than the actual volumetric reconstruction error.

Monte Carlo estimation\cite{luengo2020survey} provides a principled framework for approximating the dense volumetric objective using a sparse subset of samples while preserving consistency with the original objective. By incorporating appropriate importance weights based on the sampling distribution, the Monte Carlo estimator produces an unbiased estimate of the dense loss. Consequently, the expected value of the estimated loss and its gradients match those of the dense loss, allowing stochastic optimization to converge to the same optimum while significantly reducing computational cost. 
A naive subsampling from a non-uniform distribution leads to a biased estimator of the dense objective. To correct this, we use importance-weighted Monte Carlo estimation. The Monte Carlo estimator is defined as
\begin{align}
\hat L_{\text{MC}}
=
\frac{1}{m}
\sum_{i=1}^{m}
\frac{1}{n\,p(\mathbf{x}_i)}
\ell(\mathbf{x}_i;\Theta),
\qquad
\mathbf{x}_i \sim p(\mathbf{x}).
\end{align}

Taking expectation with respect to $p(\mathbf{x})$,
\begin{align}
\mathbb{E}[\hat L_{\text{MC}}]
=
\frac{1}{n}
\sum_{\mathbf{x}\in\Omega}
\ell(\mathbf{x};\Theta)
=
L_{\text{dense}}(\Theta),
\end{align}
showing that the estimator is unbiased. The corresponding gradient estimator is
\begin{align}
\nabla_\Theta \hat L_{\text{MC}}
=
\frac{1}{m}
\sum_{i=1}^{m}
\frac{1}{n\,p(\mathbf{x}_i)}
\nabla_\Theta \ell(\mathbf{x}_i;\Theta),
\end{align}
that satisfies $\mathbb{E}\big[\nabla_\Theta \hat L_{\text{MC}}\big] = \nabla_\Theta L_{\text{dense}}(\Theta).$

Thus, stochastic gradient descent using this estimator converges to the same optimum as dense training under standard assumptions. In practice, informative voxels often correspond to regions with high reconstruction error or strong spatial gradients. Sampling such voxels more frequently improves optimization efficiency, which naturally leads to an importance sampling formulation where the sampling probability $p(\mathbf{x})$ is proportional to a measure of voxel information.
\begin{align}
p(\mathbf{x})
=
\lambda \frac{1}{n}
+
(1-\lambda)\,p_{\text{importance}}(\mathbf{x}),
\end{align}
where $\lambda \in [0,1]$ balances uniform coverage and importance-driven sampling. The importance distribution is estimated from the reconstruction error
\begin{align}
p_{\text{importance}}(\mathbf{x})
\propto
\ell(\mathbf{x};\Theta) + \epsilon,
\end{align}
where $\epsilon > 0$ ensures non-zero probability.

This strategy prioritizes high-error regions while maintaining unbiased estimation and global coverage of the volume. Reconstruction error directly depends on the correctness of the boundary of various features in the volume and thus can be represented by high-gradient samples in the volume. A subset of voxels with high gradient magnitude are selected based on their reconstruction error, while another subset of voxels is sampled randomly from the volume. Both types of samples are periodically refreshed every few training iterations. The details of our sampling and the corresponding probability estimates are given in the supplementary material.

\subsection{Curriculum sampling strategy for structured supervision}
Our approach targets efficient shear-warp volume rendering, which composites stacks of 2D slices sampled from a 3D volume. To learn the Gaussian volumetric field while keeping training efficient, we use two complementary supervision techniques which include sparse voxel samples drawn for Monte Carlo volumetric estimation (Section~\ref{sec:mce}) and pixel samples on randomly oriented slice planes.
 We optimize the Gaussian field parameters $\Theta$ using sparse voxel coordinates $\mathbf{x}\in\Omega$ sampled from the Monte Carlo distribution $p(\mathbf{x})$ (Section~\ref{sec:mce}).
For slice-based supervision, instead of computing the loss over the entire image grid, we evaluate it only on a subset of sampled locations on the slice. In each iteration, we additionally sample a plane with arbitrary orientation, and select sparse 2D pixel locations on that plane. Each sampled pixel corresponds to a 3D coordinate $\mathbf{x}$ in the volume, and the set of samples from the same plane provides supervision on multiple points that share the same plane. Compared to isolated voxel samples, these planar samples introduce structured constraints that encourage spatial coherence within each slice plane, and thus directly improves the fidelity of the generated slices later used by shear-warp compositing. Both voxel and slice supervision update the same Gaussian volumetric function, the difference lies only in how the supervised 3D coordinates $\mathbf{x}$ are sampled and paired with ground-truth color $\mathbf{y}(\mathbf{x})$. This approach of sampling of coordinates from the slices enables the Gaussians to learn better spatial correlation while maintaining the sparsity of the input, which further makes this technique highly memory efficient.

Rather than switching supervision abruptly, we blend the two losses with weights that vary over training progress.
The weights $\lambda_v(t)$ and $\lambda_s(t)$ are scheduled over the training iteration $t$ to gradually introduce slice supervision while maintaining strong voxel supervision during early optimization. To gradually introduce planar supervision during training, we define a smooth activation schedule for the slice loss

\begin{align}
    \lambda_s(t)
=
\lambda_{s,\max}
\sigma\!\left(
\frac{t - \gamma_s}{\beta_s}
\right),
\qquad
\lambda_v(t)=1-\lambda_s(t),
\end{align}
where $\sigma(\cdot)$ denotes the sigmoid function for curriculum learning\cite{wang2019dynamic}, $\lambda_{s,\max}$ is the maximum weight assigned to the slice loss, $\gamma_s$ represents the iteration at which slice supervision becomes significant, and $\beta_s$ controls the smoothness of the transition. The voxel loss weight $\lambda_v(t)$ can be defined similarly or gradually decreased during training.

\begin{align}
\mathcal{L}(t)
=
\lambda_v(t)\,\hat L_{\text{MC}}
+
\lambda_s(t)\,L_{\text{slice}},
\end{align}
where $\mathcal{L}(t)$ denotes the total loss at iteration $t$. 
The term ${L}_{\text{slice}}$ represents the slice supervision loss computed on pixel coordinates sampled from slices. For a set of sampled slice pixels $\mathcal{S}$, the slice loss is defined as
$$
{L}_{\text{slice}}
=
\frac{1}{|\mathcal{S}|}
\sum_{\mathbf{x}\in\mathcal{S}}
\|
\mathbf{f}(\mathbf{x}) - \mathbf{y}(\mathbf{x})
\|_2^2 ,
$$
where $\mathbf{y}(\mathbf{x})$ denotes the ground-truth voxel color at $\mathbf{x}$.

Although pixels are sampled from 2D slice coordinates, each pixel corresponds to a 3D voxel location within the volume. Therefore, both voxel and slice supervision update the same Gaussian parameters through gradients of the predicted volumetric function $\mathbf{f}_\Theta$. 
This curriculum strategy combines sparse voxel supervision for global volume coverage with structured planar sampling that provides spatially correlated supervision across slices. Together, these complementary sampling strategies improve the learning of spatial coherence and geometric structures within the Gaussian volumetric representation.

\subsection{Shear–warp volume rendering using Gaussian representation}
To enable efficient real-time visualization of the learned Gaussian volumetric representation, we adopt a shear–warp volume rendering\cite{lacroute1994fast} strategy. Shear–warp rendering is chosen due to its computational efficiency for slice-based compositing and its suitability for structured 2D slice extraction from 3D volumes. Conventional shear–warp rendering takes as input a voxel grid and extracts three stacks of slices aligned with the principal axes of the volume. For an arbitrary viewing direction, a shear transformation is applied to align the viewing rays perpendicular to the selected principal stack, followed by slice-by-slice compositing. 
The ray marching~\cite{kim2021image,devkota2022deep} algorithm is a prevalent real-time rendering technique that involves sampling along every ray passing through the volume, and each ray corresponds to a pixel position on the rendered image from a camera viewing direction. The sampling operation is followed by interpolation for all the samples along all the rays for a single rendered view. These operations are computationally expensive for higher-resolution rendered views and larger volumetric data.
While efficient, this approach assumes a voxel representation which is dense by nature and requires maintaining axis-aligned slice stacks. In contrast, our approach integrates shear–warp rendering directly with the Gaussian volumetric representation. Our GPU-based renderer makes use of the fact that the Gaussians are not oriented along principal axes (unlike voxels) and therefore does not need to maintain three stacks of parallel slices but just one. We sample slice textures from the Gaussian representation along the current viewing direction, shear and combine these using the standard front-to-back alpha compositing scheme to generate an image. The stack is reused for subsequent frames as long as the camera orientation remains within a specified angular threshold. When the camera rotates beyond this threshold, the slice stack is refreshed and reoriented to match the new viewing direction. This adaptive stacking strategy reduces memory consumption by avoiding multiple precomputed stacks while maintaining rendering efficiency. 

The existing Gaussian Splatting-based renderers are mainly surface-oriented, while our goal is internal volume rendering. We therefore use shear-warp rendering, which supports slice-based compositing of volumetric interiors and is computationally cheaper than ray marching. Its slice-based memory access and incremental compositing strategy enable high rendering throughput. On the Cryosection dataset, our method achieves an average rendering speed of 43.86 FPS, compared with 15.05 FPS for ray marching achieving a  2.91$\times$ speedup. Table~\ref{tab:quantitative_organ} reports organ-wise frame rates and volume compression. These results show that the Gaussian-based shear-warp formulation enables efficient real-time rendering while preserving structural fidelity and maintaining a low memory footprint.

\section{Implementation details, results and analysis}
We evaluate our approach on volumetric datasets from multiple imaging modalities, including MRI and Cryosection volumes. For grayscale MRI data, we evaluate performance on the T2 BraTS dataset \cite{6975210} and the MRI dataset of embryonic and neonatal mouse \cite{doi:10.1073/pnas.0805747105}. The color dataset used in our experiments is the Cryosection dataset from the Visible Korean Project \cite{park2005visible}. We further evaluate our approach on multiple organs from this dataset, including lungs, liver, heart, and brain, where the organ-level volumes are extracted using the provided segmentation masks. The ground-truth supervision consists of a sparse voxel set sampled from the dense voxel grid using a hybrid strategy based on gradient magnitude and random sampling. We select the sparse voxel set such that the total memory footprint of the Gaussian representations remains within 25\% of the original volumetric data size. Within this sparse set, 60\% of voxels are selected from high gradient magnitude regions, while the remaining 40\% are randomly sampled across the volume. Optimization is performed using the Adam optimizer with a learning rate of $5 \times 10^{-4}$, and the Gaussian parameters are trained for 120 epochs. We evaluate reconstruction accuracy by comparing the Gaussian-reconstructed volume with the original dense voxel grid. Rendering quality is evaluated by comparing shear–warp rendered images generated from the Gaussian representation with rendered images from the dense voxel grid. The reconstruction accuracy and rendering quality are analyzed for both the reconstructed voxel grid and arbitrary rendered views obtained from the Gaussian representation. 

\subsection{Comparisons}

\subsubsection{Datasets}

\begin{figure}[t]
\centering
\renewcommand{\arraystretch}{1.3}

\begin{tabular}{c c c c c c c}
\raisebox{0.24in}{\rotatebox[origin=c]{90}{\scriptsize\shortstack{\renewcommand{\arraystretch}{0.9} GT}}} &
\img{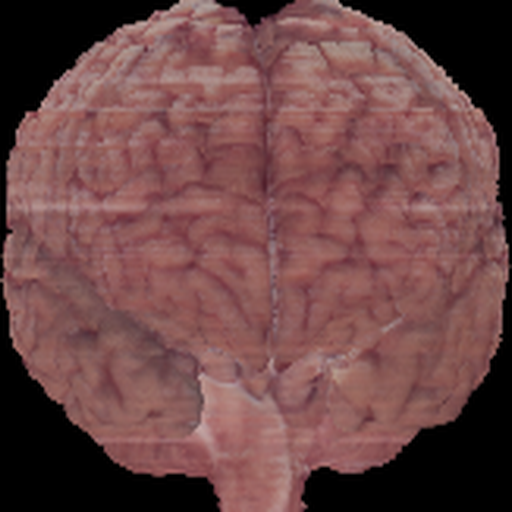} &
\img{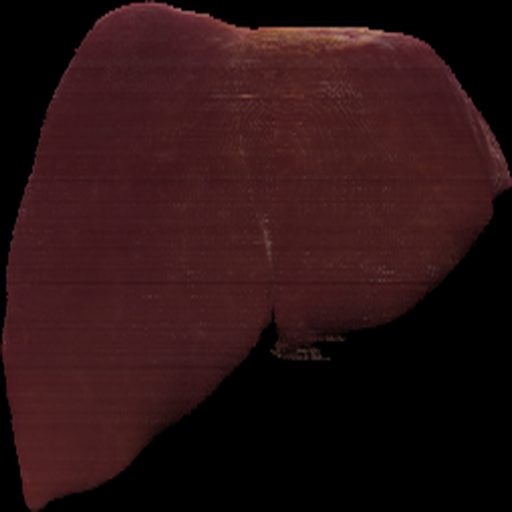} &
\img{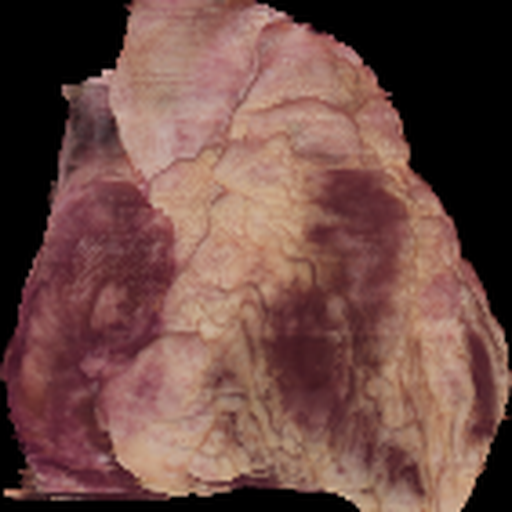} &
\img{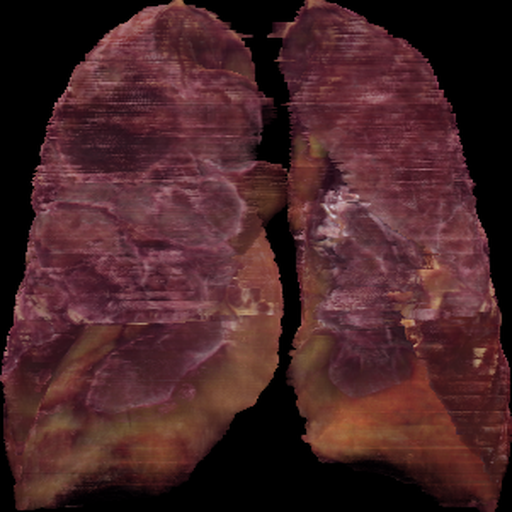} &
\img{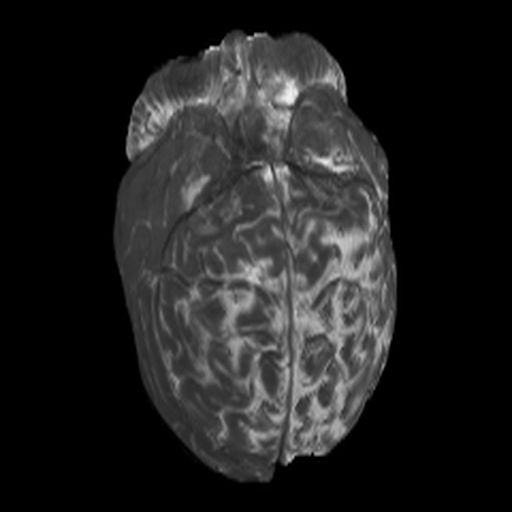} &

\img{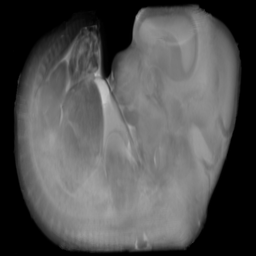} \\

\raisebox{0.24in}{\rotatebox[origin=c]{90}{\scriptsize\shortstack{\renewcommand{\arraystretch}{0.9} Pred.}}} &
\img{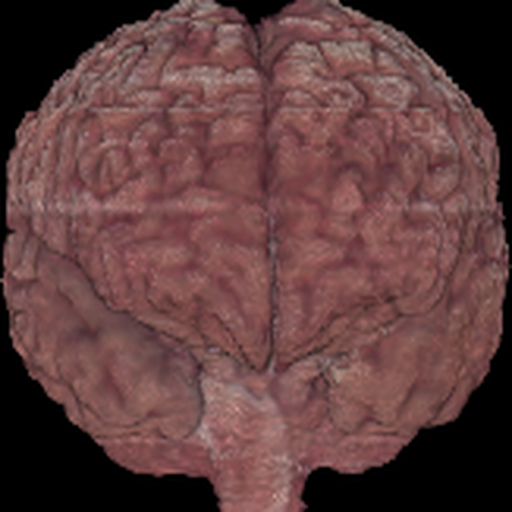} &
\img{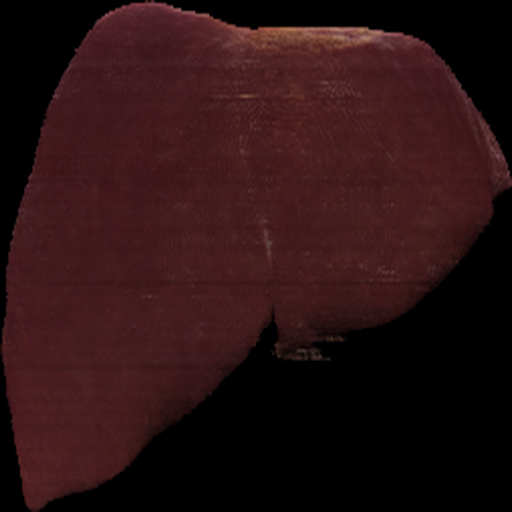} &
\img{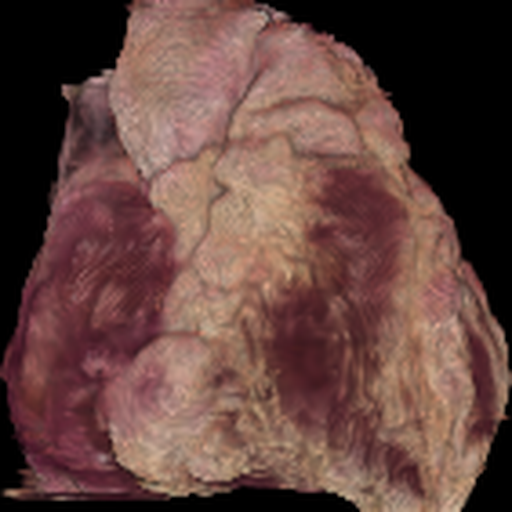} &
\img{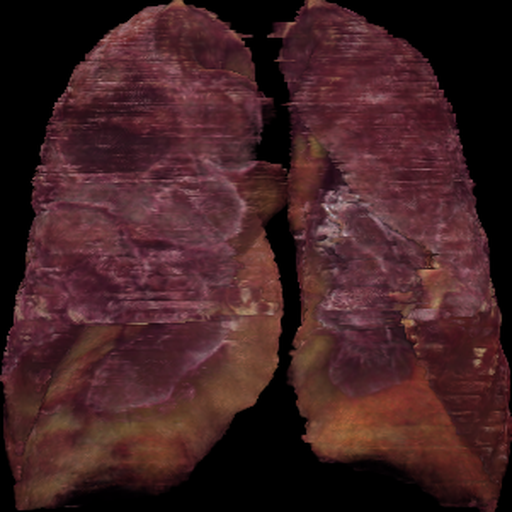} &
\img{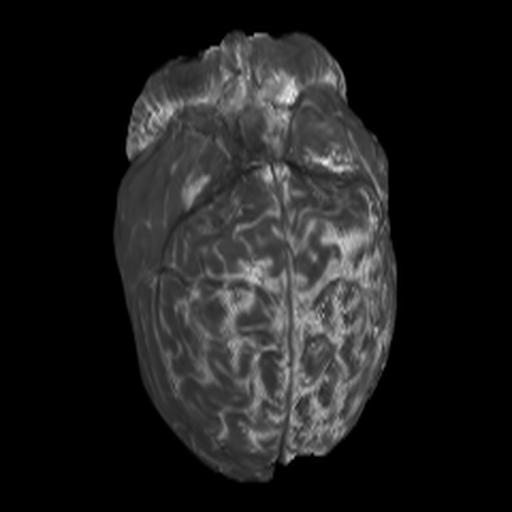} &

\img{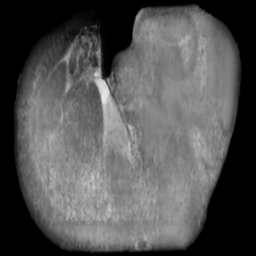} \\

& {\scriptsize\shortstack{Brain \\ Cryo\cite{park2005visible}}} &
  {\scriptsize\shortstack{Liver \\Cryo\cite{park2005visible}}} &
  {\scriptsize\shortstack{Heart \\ Cryo\cite{park2005visible}}} &
  {\scriptsize\shortstack{Lungs \\ Cryo\cite{park2005visible}}} &
  {\scriptsize\shortstack{BraTs \\ MRI\cite{6975210}}} &
  {\scriptsize\shortstack{Mouse \\ MRI\cite{doi:10.1073/pnas.0805747105}}} \\

\end{tabular}

\caption{Visual comparison of rendered views for different organs, including brain, liver, heart, and lungs from Cryosection data, BraTS MRI data, and mouse neonatal MRI data.}
\label{tab:organs_qual}

\end{figure}

The evaluation metrics are computed for grayscale MRI modalities, including the BraTS\cite{6975210} dataset and the embryonic–neonatal mouse MRI dataset. The RGB modality corresponds to the Cryosection dataset, along with organ-level volumes including lungs, liver, heart, and brain.
The qualitative analysis of our approach on the BraTS dataset is shown in  Fig.~\ref{tab:organs_qual}. The visual results illustrate the dense volume reconstructed from the sparse Gaussian representation. It can be observed that the proposed technique successfully generates rendered views using the shear–warp rendering algorithm from the Gaussian-based volumetric representation. Furthermore, the method efficiently preserves structural details in both the rendered views and volumetric cross-sectional slices across all modalities. Performance is quantified using Mean Squared Error (MSE)\cite{sara2019image}, Peak Signal-to-Noise Ratio (PSNR)\cite{sara2019image}, Structural Similarity Index (SSIM)\cite{zhu2017unpaired}, and Multi-Scale SSIM (MS-SSIM)\cite{zhu2017unpaired} as shown in Table~\ref{tab:quantitative_organ}. The results demonstrate that our technique consistently achieves low MSE values and high PSNR, SSIM, and MS-SSIM scores across all modalities, indicating that the sparse Gaussian representation effectively captures the structural and textural characteristics of the target volumes.
\subsubsection{Sparse representations}
\begin{figure}[!htp]
\centering

\makebox[\linewidth][c]{%
\begin{tabular}{cccccccccc}

\raisebox{0.24in}{\rotatebox[origin=c]{90}{\scriptsize\shortstack{\renewcommand{\arraystretch}{0.9}Cross-\\section}}} &
\img{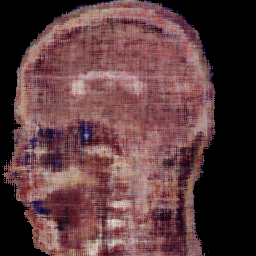} &
\img{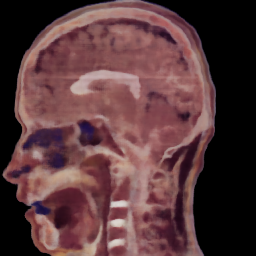} &
\img{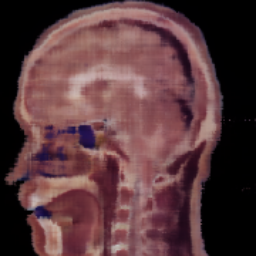} &
\img{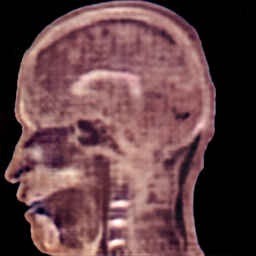} &
\img{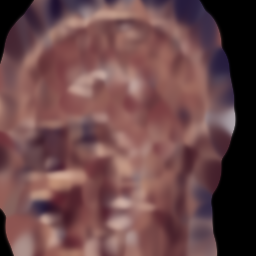} &
\img{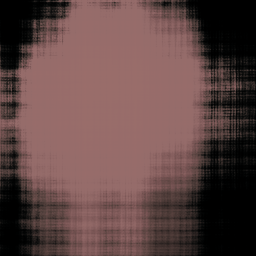} &
\img{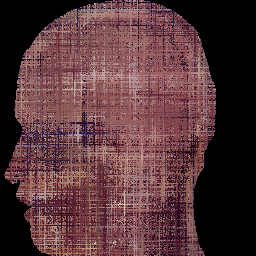} &
\img{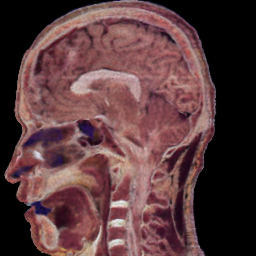} &
\img{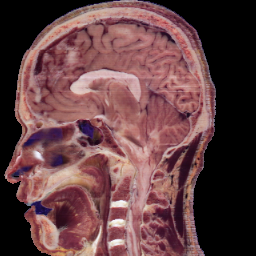} \\

\raisebox{0.24in}{\rotatebox[origin=c]{90}{\scriptsize\shortstack{\renewcommand{\arraystretch}{0.9}Cross-\\section}}} &
\img{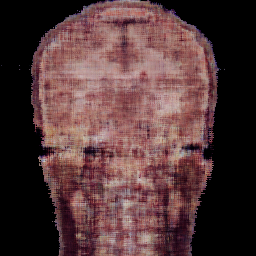} &
\img{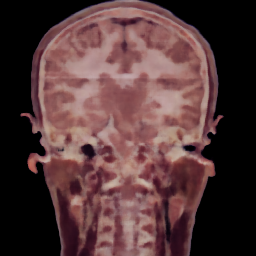} &
\img{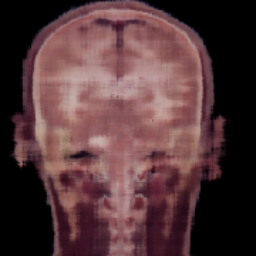} &
\img{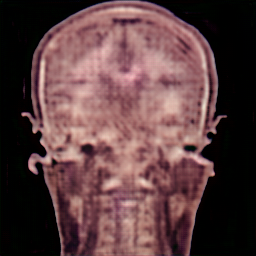} &
\img{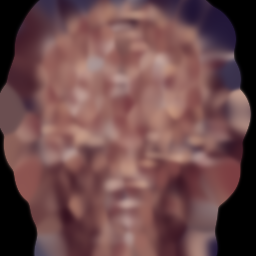} &
\img{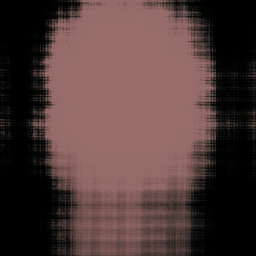} &
\img{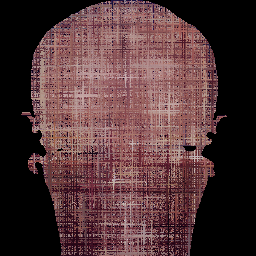} &
\img{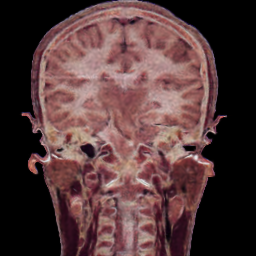} &
\img{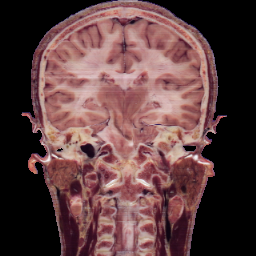} \\

\raisebox{0.24in}{\rotatebox[origin=c]{90}{\scriptsize\shortstack{\renewcommand{\arraystretch}{0.9}Rendered\\view}}} &
\img{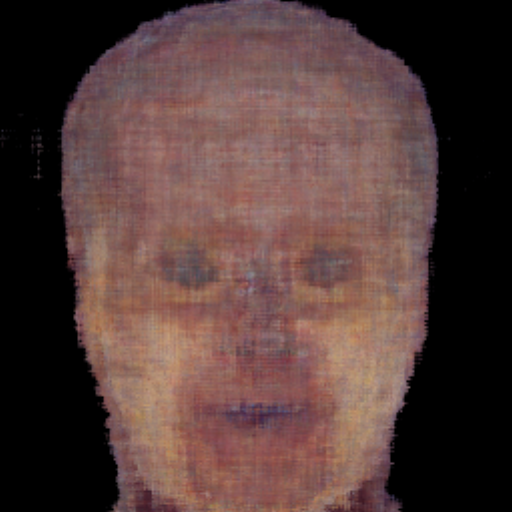} &
\img{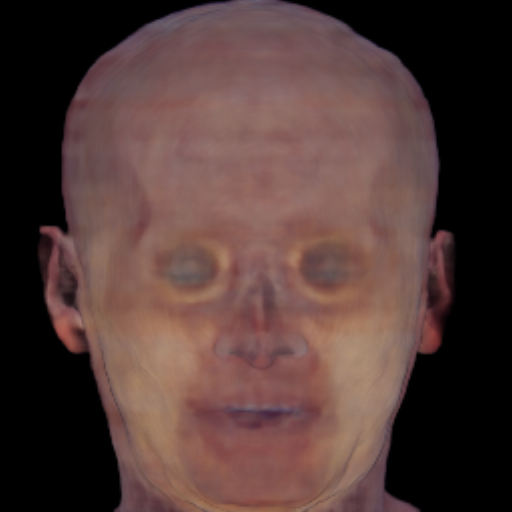} &
\img{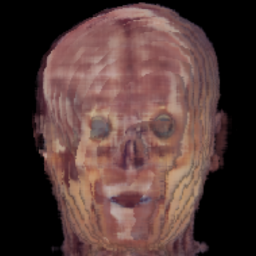} &
\img{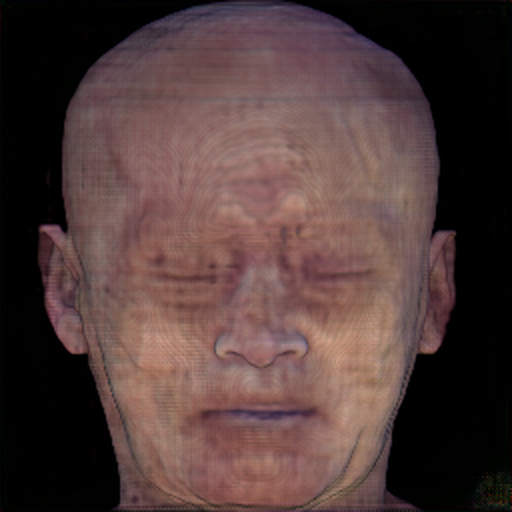} &
\img{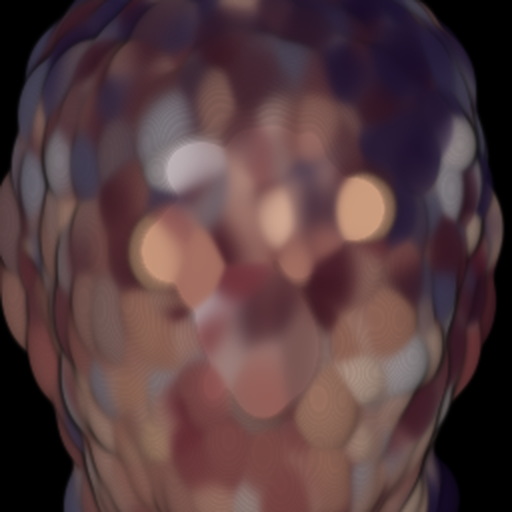} &
\img{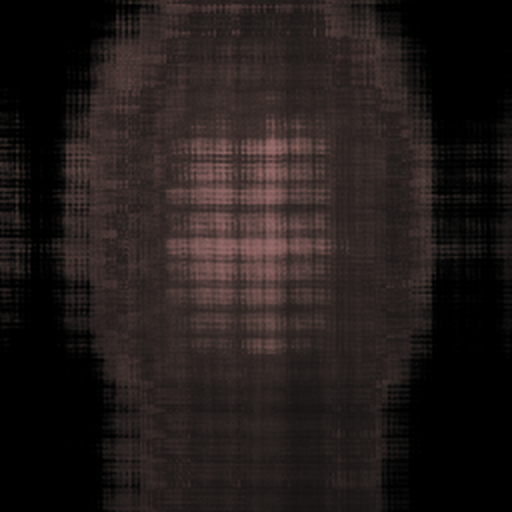} &
\img{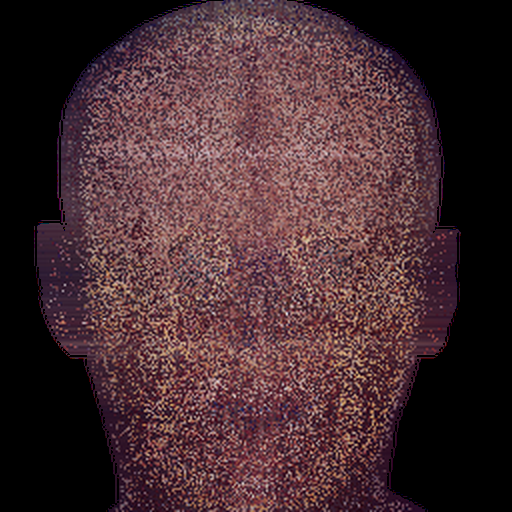} &
\img{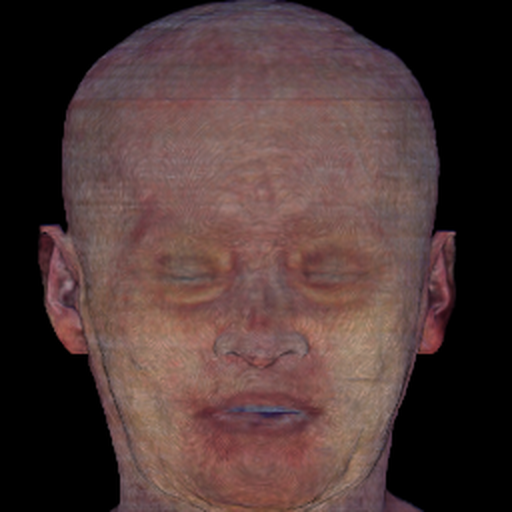} &
\img{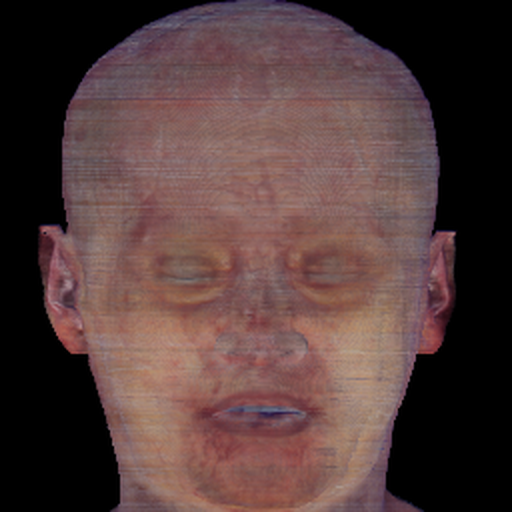} \\

& {\scriptsize\shortstack{Voxel \\ MLP\cite{mildenhall2021nerf}}} &
  {\scriptsize\shortstack{Slice \\ MLP\cite{mildenhall2021nerf}}} &
  {\scriptsize\shortstack{Trans \\ former\cite{wang2021multi}}} &
  {\scriptsize UNet\cite{cciccek20163d}} &
  {\scriptsize\shortstack{Free \\ Splatter\cite{xu2025freesplatter}}} &
  {\scriptsize\shortstack{CVT-\\xRF\cite{zhong2024cvt}}} &
  {\scriptsize\shortstack{GNT-\\MOVE\cite{cong2023enhancing}}} &
  {\scriptsize Ours} &
  {\scriptsize GT} \\

\end{tabular}%
}

\caption{Comparisons with learning-based volumetric representation methods.}
\label{tab:compar_qual}
\end{figure}

We compare our approach with several high-capacity deep learning models used for learning complex data patterns, including MLPs \cite{mildenhall2021nerf}, Transformers \cite{wang2021multi}, and UNet \cite{cciccek20163d}. These models rely on large numbers of learnable parameters and are evaluated for reconstructing dense voxel grids from sparse voxel observations. As shown in Fig.~\ref{tab:compar_qual}, we evaluate MLP-based representations using two supervision strategies: voxel-based and slice-based supervision. In voxel supervision, the MLP maps sparse voxel coordinates sampled from the dense volume to their corresponding color intensities in the Cryosection data. In slice supervision, sparse voxel coordinates are sampled from cross-sectional planes and mapped to the color intensities of those planes. In both settings, the MLP fails to preserve fine structural details and struggles to reconstruct high-quality rendered views. For the Transformer-based\cite{wang2021multi} representation, we train the model using sparse $8 \times 8 \times 8$ 3D patches sampled from the volume. However, the Transformer struggles to preserve fine anatomical structures and introduces synthetic artifacts in regions with weak anatomical signals. For the UNet-based approach \cite{cciccek20163d}, we use a local 3D voxel neighborhood of size $8 \times 8 \times 8$ and predict voxel intensities using 3D convolutional layers. While this captures local context, 3D convolutions increase computational overhead, and the model underperforms in capturing complex color variations and anatomical structures.
We also include GNT-MOVE~\cite{cong2023enhancing}, CVT-xRF~\cite{zhong2024cvt}, and FreeSplatter~\cite{xu2025freesplatter} as recent Transformer, NeRF, and Gaussian Splatting-based reconstruction methods. These approaches are representative of modern high-capacity rendering architectures that use attention mechanisms, radiance-field modeling, or Gaussian-based scene representations to reconstruct complex visual data. Since these methods were not originally designed for sparse volumetric reconstruction of dense Cryosection data, we adapt their input representations, training hyperparameters, and minor architectural settings to fit our problem while keeping their core model designs unchanged. As shown in Fig.~\ref{tab:compar_qual}, they do not consistently preserve fine internal anatomical structures or dense color variations in the Cryosection volume, whereas our approach directly fits a compact volumetric representation for internal 3D visualization.
We further evaluate reconstruction and rendering accuracy using PSNR, MSE, SSIM, and MS-SSIM, as shown in Table~\ref{tab:quantitative_efficiency_comparison}. Our Gaussian-based representation consistently achieves lower MSE and higher PSNR, SSIM, and MS-SSIM than the deep learning baselines, indicating better preservation of structural and textural details. The table also reports efficiency in terms of training time, memory usage, FLOPs, and FPS, which are critical for real-time volumetric visualization. Compared with neural baselines, our method achieves high visual fidelity with substantially lower computational cost, demonstrating that the sparse Gaussian representation supports efficient reconstruction and real-time rendering. $^\dagger$FPS is not reported for GNT-MOVE~\cite{cong2023enhancing} because its rendering pipeline requires view-dependent neural inference with a large number of parameters, making real-time rendering infeasible under our evaluation setup.
\begin{table}[!t]
\centering
\caption{Quantitative analysis on different datasets.}
\scriptsize
\begin{tabular}{lccccr}
\toprule
\textbf{Dataset} 
& \textbf{PSNR}
& \textbf{SSIM}
& \textbf{MS-SSIM}
& \textbf{FPS}
& \textbf{Compression ratio} \\
\midrule
Brain\cite{park2005visible}           & 32.10  & 0.943 & 0.991 & 45.5 & 9.34{:1}\\
Liver\cite{park2005visible}           & 43.16  & 0.981 & 0.996 & 50.18 & 9.45{:1} \\
Heart\cite{park2005visible}           & 32.19 & 0.958 & 0.990 & 48.57 & 10.09{:1} \\
Lungs\cite{park2005visible}           & 34.93 & 0.952 & 0.989 & 39.77 & 4.73{:1} \\
BraTS\cite{6975210}           & 40.88  & 0.988 & 0.998 & 44.82 & 11.31{:1} \\
Mouse Neonatal\cite{doi:10.1073/pnas.0805747105}  & 33.48 & 0.956 & 0.987 & 46.56 & 9.94{:1} \\
\bottomrule
\end{tabular}
\label{tab:quantitative_organ}
\end{table}

\begin{table}[!htp]
\centering
\caption{Quantitative comparison with baselines. Top results are highlighted as \bestcap{1st}, \secondcap{2nd}, and \thirdcap{3rd}.}
\label{tab:quantitative_efficiency_comparison}
\scriptsize
\setlength{\tabcolsep}{0.45pt}
\renewcommand{\arraystretch}{0.80}
\resizebox{\columnwidth}{!}{%
\begin{tabular}{@{}lcccc|cccc|ccccc@{}}
\toprule
\textbf{Method}
& \multicolumn{4}{c|}{\textbf{Reconstruction}}
& \multicolumn{4}{c|}{\textbf{Rendering}}
& \multicolumn{5}{c}{\textbf{Efficiency}} \\
\cmidrule(lr){2-5}
\cmidrule(lr){6-9}
\cmidrule(lr){10-14}
& \textbf{PSNR}$\uparrow$
& \textbf{MSE}$\downarrow$
& \textbf{SSIM}$\uparrow$
& \textbf{MS-S}
& \textbf{PSNR}$\uparrow$
& \textbf{MSE}$\downarrow$
& \textbf{SSIM}$\uparrow$
& \textbf{MS-S}
& \textbf{Epochs}
& \textbf{Time}
& \textbf{Mem}
& \textbf{FLOPs}
& \textbf{FPS} \\
&
&
&
&\textbf{SIM}$\uparrow$
&
&
&
&\textbf{SIM}$\uparrow$
&
&(hours)
&(GB)
&
&\\
\midrule

Voxel MLP~\cite{mildenhall2021nerf}
& 23.63 & \second{0.000434} & 0.772 & 0.871
& \second{46.610} & \second{0.000176} & 0.968 & 0.980
& \second{200} & \third{6.7} & 15 & $9.50{\times}10^{12}$ & 18.99 \\

Slice MLP~\cite{mildenhall2021nerf}
& \second{30.02} & \third{0.000995} & \second{0.920} & \second{0.969}
& 36.32 & \third{0.000248} & \third{0.973} & \third{0.983}
& \third{300} & 35.0 & \third{14} & $1.88{\times}10^{13}$ & 14.27 \\

Transformer~\cite{wang2021multi}
& 9.60 & 0.052000 & 0.631 & 0.404
& 29.73 & 0.014900 & 0.873 & 0.481
& \second{200} & 33.3 & 30 & $1.91{\times}10^{15}$ & 0.11 \\

UNet~\cite{cciccek20163d}
& \third{26.70} & 0.002130 & \third{0.798} & \third{0.943}
& \third{42.10} & 0.000520 & \second{0.982} & \second{0.985}
& \second{200} & 16.7 & 36 & \third{$2.05{\times}10^{12}$} & 15.00 \\

CVT-xRF~\cite{zhong2024cvt}
& 18.98 & 0.012657 & 0.591 & 0.560
& 15.09 & 0.033922 & 0.536 & 0.551
& 500 & 19.3 & 22 & $2.84{\times}10^{15}$ & \third{30.17} \\

FreeSplatter~\cite{xu2025freesplatter}
& 14.74 & 0.033608 & 0.483 & 0.515
& 16.24 & 0.023822 & 0.569 & 0.426
& 700 & \best{0.13} & \best{11} & \second{$1.27{\times}10^{12}$} & \second{36.01} \\

GNT-MOVE~\cite{cong2023enhancing}
& 11.28 & 0.074417 & 0.046 & 0.317
& 15.11 & 0.030866 & 0.084 & 0.413
& \third{300} & 39.2 & \second{12} & $1.05{\times}10^{15}$ & N/A$^\dagger$ \\

\textbf{Ours}
& \best{34.81} & \best{0.000398} & \best{0.956} & \best{0.993}
& \best{54.87} & \best{0.000021} & \best{0.995} & \best{0.993}
& \best{120} & \second{\textbf{6.0}} & 28 & \best{$9.17{\times}10^{9}$} & \best{43.86} \\

\bottomrule
\end{tabular}%
}

\end{table}

\subsection{Ablation study}

\begin{figure}[!htp]
\centering
\begingroup
\setlength{\tabcolsep}{2pt}
\renewcommand{\arraystretch}{1.15}

\begin{tabular*}{\linewidth}{@{\extracolsep{\fill}} >{\raggedleft\arraybackslash}p{\labw} c c c c c c c @{}}

\raisebox{0.24in}{\rotatebox[origin=c]{90}{\scriptsize\shortstack{\renewcommand{\arraystretch}{0.9}Cross-\\section}}} &
\includegraphics[width=\imgw]{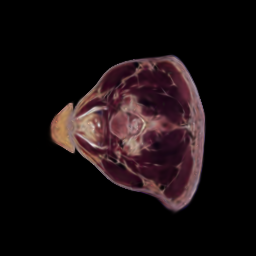} &
\includegraphics[width=\imgw]{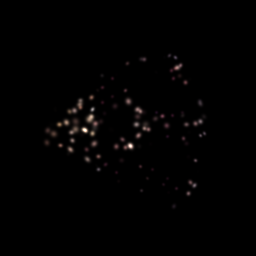} &
\includegraphics[width=\imgw]{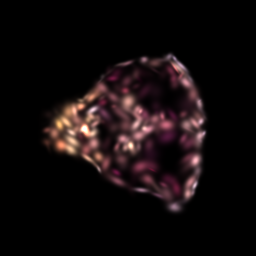} &
\includegraphics[width=\imgw]{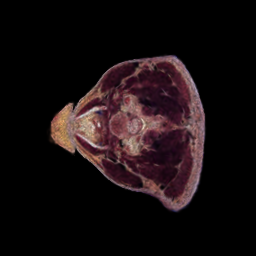} &
\includegraphics[width=\imgw]{Images/ablations/MCE_P1/slices/pred_axial_d0255.png} &
\includegraphics[width=\imgw]{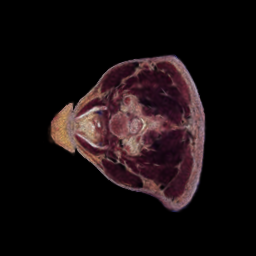} &
\includegraphics[width=\imgw]{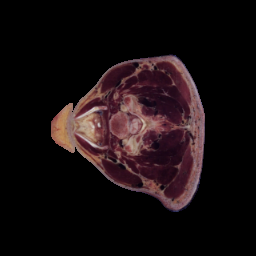} \\

\raisebox{0.24in}{\rotatebox[origin=c]{90}{\scriptsize\shortstack{\renewcommand{\arraystretch}{0.9}Rendered\\view}}} &
\includegraphics[width=\imgw]{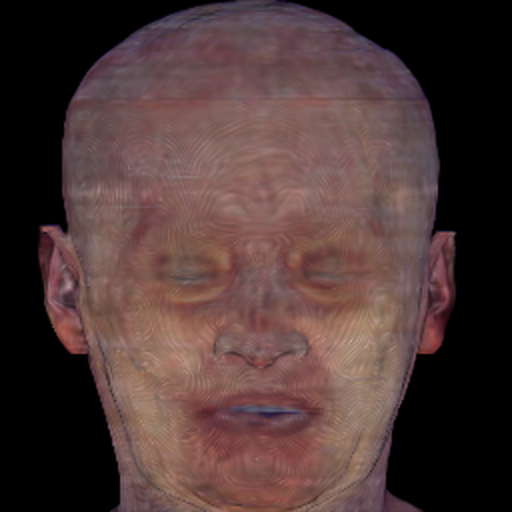} &
\includegraphics[width=\imgw]{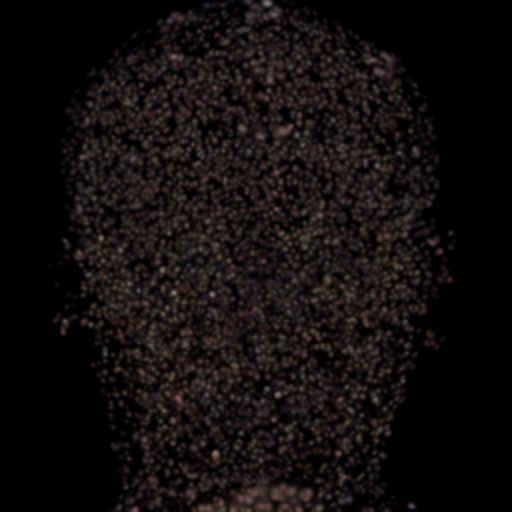} &
\includegraphics[width=\imgw]{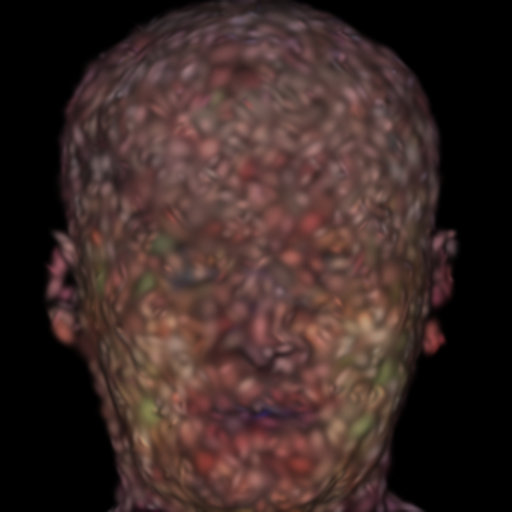} &
\includegraphics[width=\imgw]{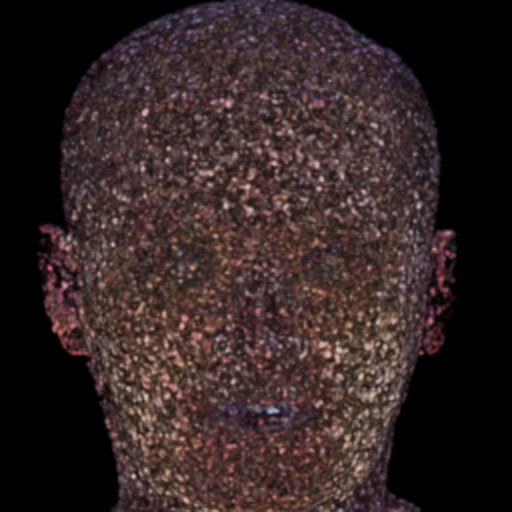} &
\includegraphics[width=\imgw]{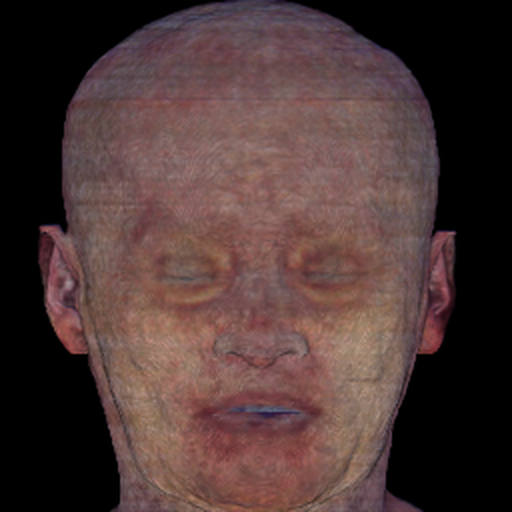} &
\includegraphics[width=\imgw]{Images/ablations/MCE_P2/principal_views_alpha08/view_x.png} &
\includegraphics[width=\imgw]{Images/ablations/gt/principal_views_alpha08/view_x.png} \\

\raisebox{0.24in}{\rotatebox[origin=c]{90}{\scriptsize\shortstack{\renewcommand{\arraystretch}{0.9}Rendered\\view}}} &
\includegraphics[width=\imgw]{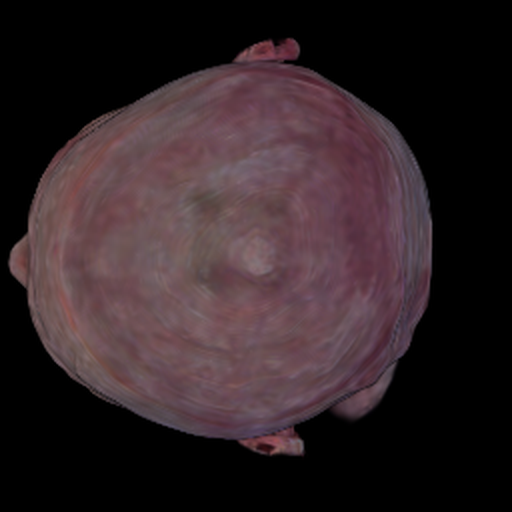} &
\includegraphics[width=\imgw]{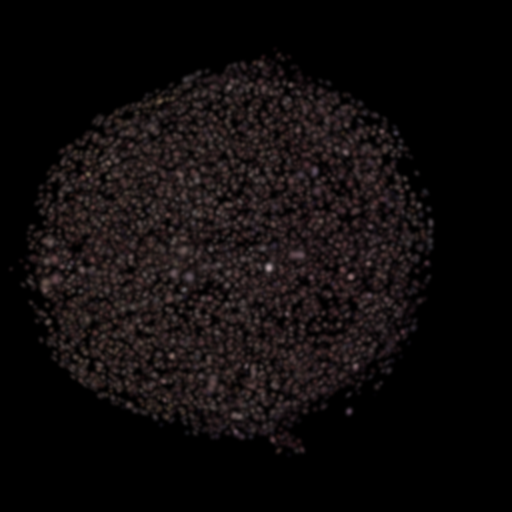} &
\includegraphics[width=\imgw]{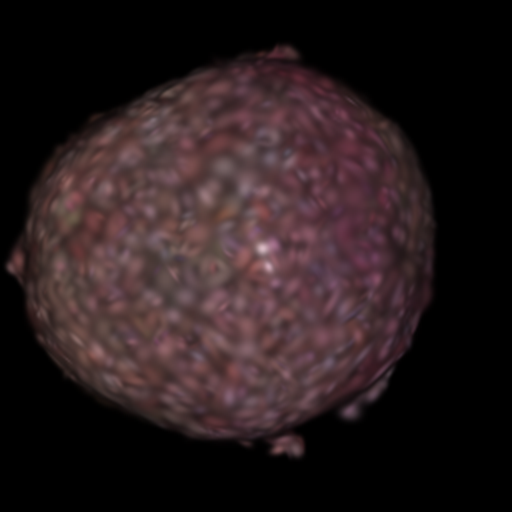} &
\includegraphics[width=\imgw]{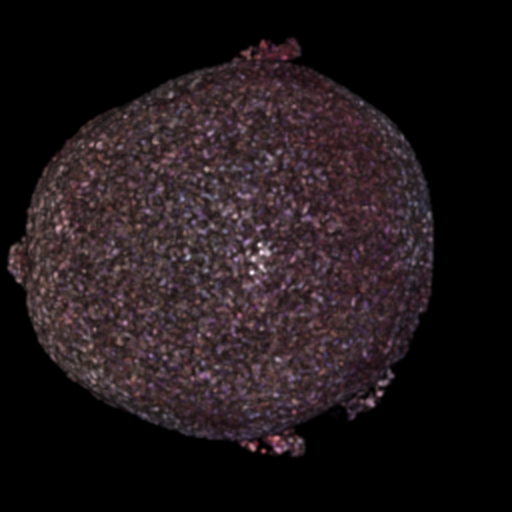} &
\includegraphics[width=\imgw]{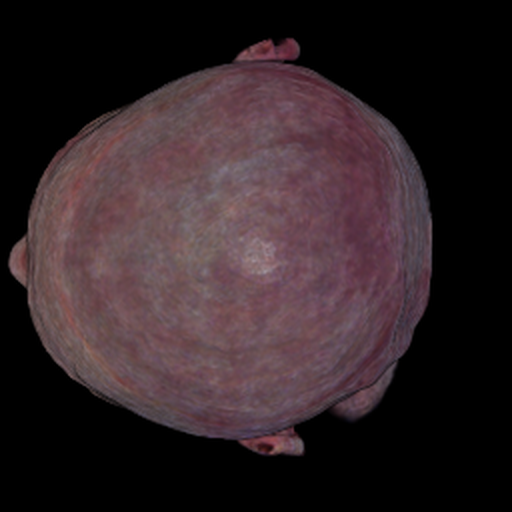} &
\includegraphics[width=\imgw]{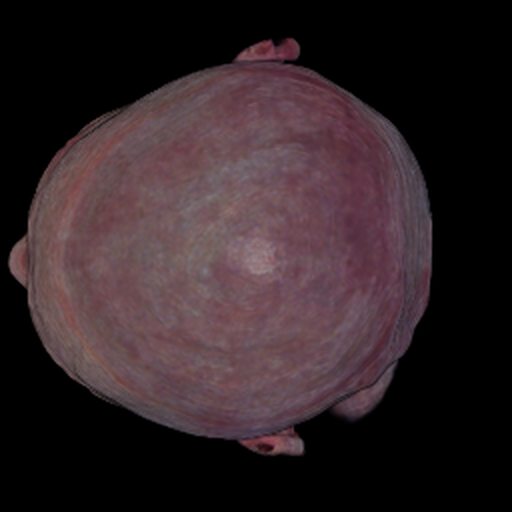} &
\includegraphics[width=\imgw]{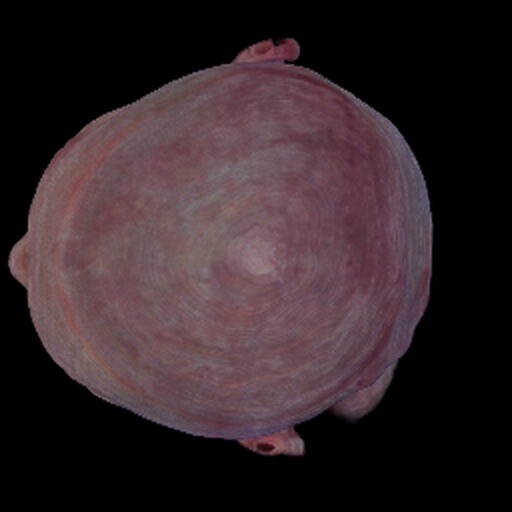} \\

& \textbf{A$_{1}$} & \textbf{A$_{2}$} & \textbf{A$_{3}$} & \textbf{A$_{4}$} & \textbf{A$_{5}$} & \textbf{Ours} & \textbf{GT} \\

\end{tabular*}
\endgroup
\caption{Visual comparisons of the ablation study on cross sections and rendered views from the shear-warp renderer.} 
\label{tab:image_table}
\end{figure}

We perform an ablation study to analyze the contribution of the proposed sampling strategy and curriculum learning–based supervision and evaluate reconstruction quality using PSNR, MSE, SSIM, and MS-SSIM.

\noindent\textbf{A$_{1}$(Random voxel supervision):}
In this experiment, we optimize the Gaussian representation using randomly sampled voxel coordinates together with high-gradient-magnitude voxels. At each iteration, a new set of voxel samples is selected, and the reconstruction loss is computed against the dense volume. In contrast, our approach initializes Gaussians over a sparse voxel set and approximates the dense volume using MCE. Our method captures finer structural details in both reconstructed volumes and rendered views. The evaluation metrics further confirm improved reconstruction under sparse voxel supervision.

\noindent\textbf{A$_{2}$(Curriculum learning with random voxel then random slices):} 
This setting introduces slice supervision together with voxel supervision under random sampling during Gaussian optimization. While slice supervision progressively incorporates finer structural details through curriculum learning, random voxel sampling leads to unstable global approximation and further destabilizes the learning of Gaussian parameters.

\noindent\textbf{A$_{3}$ (Curriculum learning random slices then random voxel):}
In this ablation, slice supervision is applied first, followed by voxel supervision under random sampling with dense approximation. Since slice supervision is introduced before stable global voxel constraints are established, the Gaussian parameters are initially optimized using limited planar information. This results in unstable learning of the volumetric structure, and the later introduction of random voxels does not sufficiently correct the global approximation.

\noindent\textbf{A$_{4}$ (Only slice supervision):}
This configuration uses only slice supervision without voxel supervision. Although slices provide structured planar constraints and capture local spatial continuity, they cover limited regions of the volume at each iteration. Without sparse voxel supervision across the volume, the Gaussian representation lacks global spatial guidance. This limits accurate reconstruction of the full volumetric structure.

\noindent\textbf{A$_{5}$ (MCE voxel based supervision)}:
In this setting, MCE is used for voxel sampling while optimizing Gaussian parameters using voxel supervision only. Importance-weighted sampling improves coverage of informative regions while remaining consistent with the dense reconstruction objective. This updates Gaussian parameters using more representative voxel samples and stabilizes optimization. However, without slice supervision, the model lacks structured planar constraints and shows lower performance in rendered views than our method.

\begin{table}[t]
\centering
\caption{Quantitative results of ablation on reconstruction and rendering quality.}
\scriptsize
\begin{tabular}{lccccccc}
\toprule
\textbf{Method} & \multicolumn{3}{c}{\textbf{Reconstruction quality}} & \multicolumn{3}{c}{\textbf{Rendering quality}}\\
 \cmidrule(lr){2-4}
 \cmidrule(lr){5-7} 
& \textbf{PSNR}$\uparrow$ 
& \textbf{SSIM}$\uparrow$ 
& \textbf{MS-SSIM}$\uparrow$
& \textbf{PSNR}$\uparrow$ 
& \textbf{SSIM}$\uparrow$ 
& \textbf{MS-SSIM}$\uparrow$ \\
\midrule
\textbf{A$_{1}$} & 33.46  & 0.956   & 0.990 & 50.7 & 0.962   & 0.964 \\ 
\textbf{A$_{2}$} & 12.16   & 0.060 & 0.214 & 27.10  & 0.826 & 0.789\\ 
\textbf{A$_{3}$} & 22.64  & 0.727 & 0.850 & 40.59   & 0.932  & 0.937\\ 
\textbf{A$_{4}$} & 14.350 & 0.646 & 0.512 & 33.18  & 0.878 & 0.887\\ 
\textbf{A$_{5}$} & 32.40 & 0.943  & 0.988 & 52.34  & 0.968  & 0.991\\ 
\textbf{Ours} & \textbf{34.81} & \textbf{0.956} & \textbf{0.993} & \textbf{54.870}  & \textbf{0.995} & \textbf{0.993}\\
\bottomrule
\end{tabular}
\label{tab:quantitative_A}
\end{table}


\section{Limitations}
The proposed representation has limitations due to the smooth nature of Gaussian primitives. Since each primitive models the signal using a smooth spatial kernel, the reconstructed volumetric field may exhibit slight blurring near sharp boundaries or high-frequency details. This behavior is inherent to Gaussian-based representations that approximate volumetric signals using smooth basis functions. Additionally, small bright dot-like artifacts in certain regions of the reconstructed volume appear occasionally. These artifacts likely arise from the highly sparse supervision used during optimization, where only a limited subset of voxel samples guides the Gaussian parameter updates. Although the current framework produces visually consistent reconstructions even under strong sparsity constraints (as shown in the supplementary material), improved sampling strategies or regularization could further reduce these artifacts in future work.
\section{Conclusion}

In this work, we present a Gaussian-based volumetric representation for efficient reconstruction and visualization of high-resolution medical volumes. The volume is modeled as a continuous Gaussian field that can be evaluated at arbitrary spatial coordinates, enabling dense reconstruction from sparse voxel supervision. To optimize this representation, we use Monte Carlo volumetric estimation to approximate the dense reconstruction objective using a sparse voxel subset while remaining consistent with the full objective.We further introduce a curriculum learning–based sampling strategy that combines sparse voxel samples with planar slice samples during training. While voxel samples provide global volumetric coverage, slice supervision introduces structured spatial constraints that improve the learning of anatomical structures. The learned Gaussian representation supports efficient shear–warp volume rendering for high-quality visualization of volumetric medical data across modalities such as MRI and Cryosection. The effectiveness of our approach is demonstrated across multiple modalities, compared with learning-based volumetric reconstruction methods, and validated through comprehensive ablation studies.
\section*{Acknowledgements}
This research was supported by the Science and Engineering Research Board (SERB) of the Department of Science and Technology (DST) of India (Grant No. CRG/2020/005792). The Visible Korean Human dataset for our research was provided by the Korea Institute of Science and Technology Information (KISTI), South Korea.

%
%
\bibliographystyle{splncs04}
\bibliography{main}
\end{document}